%% file: objective2015nips.tex
\documentclass{article}
\usepackage{nips15submit_e}

\input{preamble.tex}

\input{preamble_math.tex}
\input{preamble_acronyms}

\newcommand{\bz}{\mathbf{z}}
\newcommand{\bx}{\mathbf{x}}
\newcommand{\bB}{\mathbf{B}}
\newcommand{\bb}{\mathbf{b}}
\newcommand{\by}{\mathbf{y}}

\nipsfinalcopy % uncomment only at the end

\title{Objective Variables for Probabilistic Revenue Maximization in
  Second-Price Auctions with Reserve}

\author{%
Maja R. Rudolph                                 \\
Department of Computer Science              \\
Columbia University                         \\
\texttt{mrr2163@columbia.edu}               \\
\And
Joseph G. Ellis				\\
Department of Electrical Engineering              \\
Columbia University                         \\
\texttt{jge2105@columbia.edu}               \\
\And
David M.~Blei                               \\
Data Science Institute                      \\
Departments of Computer Science, Statistics \\
Columbia University                         \\
\texttt{david.blei@columbia.edu}            \\
}

\begin{document}
\maketitle

\begin{abstract}
  Many online companies sell advertisement space in second-price auctions with reserve.  In this paper, we develop a
  probabilistic method to learn a profitable strategy to set the
  reserve price.  We use historical auction data with features to fit
  a predictor of the best reserve price.  This problem is
  delicate---the structure of the auction is such that a reserve price
  set too high is much worse than a reserve price set too low.  To
  address this we develop \textit{objective variables}, a new
  framework for combining probabilistic modeling with optimal
  decision-making.  
Objective variables are "hallucinated observations" that transform the revenue maximization task into a regularized maximum likelihood estimation problem, which we solve with an EM algorithm.
This framework enables a variety of prediction mechanisms to set the reserve price.
As examples, we study objective variable methods with regression, kernelized regression, and neural networks on simulated and real data.
Our methods outperform previous approaches both in terms of scalability and profit.
\end{abstract}

\section{Introduction}

Many online companies earn money from auctions, selling advertisement
space or other items. One widely used auction paradigm is second-price
auctions with reserve \citep{easley}. In this paradigm, the company
sets a {\em reserve price}, the minimal price at which they are
willing to sell, before potential buyers cast their bids. If the
highest bid is smaller than the reserve price then there is no
transaction; the company does not earn money. If any bid is larger
than the reserve price then the highest bidding buyer wins the
auction, and the buyer pays the larger of the second highest bid and
the reserve price.  To maximize their profit from a specific auction,
the host company wants to set the reserve price as close as possible
to the (future, unknown) highest bid, but no higher.

Imagine a company which hosts second-price auctions with reserve to
sell baseball cards.  This auction mechanism is designed to be
incentive compatible \citep{bar2002incentive}, which means that it is
advantageous for baseball enthusiasts to bid exactly what they are
willing to pay for the Stanley Kofax baseball card they are eager to
own\footnote{In contrast, the auction mechanism used on eBay is not
  incentive compatible since the bids are not sealed. As a result,
  experienced bidders refrain from bidding the true amount they are
  willing to pay until seconds before the auction ends to keep sale
  prices low.}.  Before each auction starts the company has to set the
reserve price.  When companies run millions of auctions of similar
items, they have the opportunity to learn how to opportunistically set
the reserve price from their historical data.  In other words, they
can try to learn their users' value of different items, and take
advantage of this knowledge to maximize profit.  This is the problem
that we address in this paper.

We develop a probabilistic model that predicts a good reserve price
from prior features of an auction.  These features might be properties
of the product, such as the placement of the advertisement, properties
of the potential buyers, such as each one's average past bids, or
other external features, such as time of day of the auction.  Given a
data set of auction features and bids, our method learns a predictor
of reserve price that maximizes the profit of future auctions.

\begin{figure}[t]
\centering
\subcaptionbox
{Revenue function of $4$ auctions 
    from the eBay data set as a function of reserve price. In
    second-price auctions with reserve the revenue depends on
    the highest and the second highest bid (dashed lines).\label{revFun}}[0.49\textwidth]
{\includegraphics[width=0.49\textwidth]{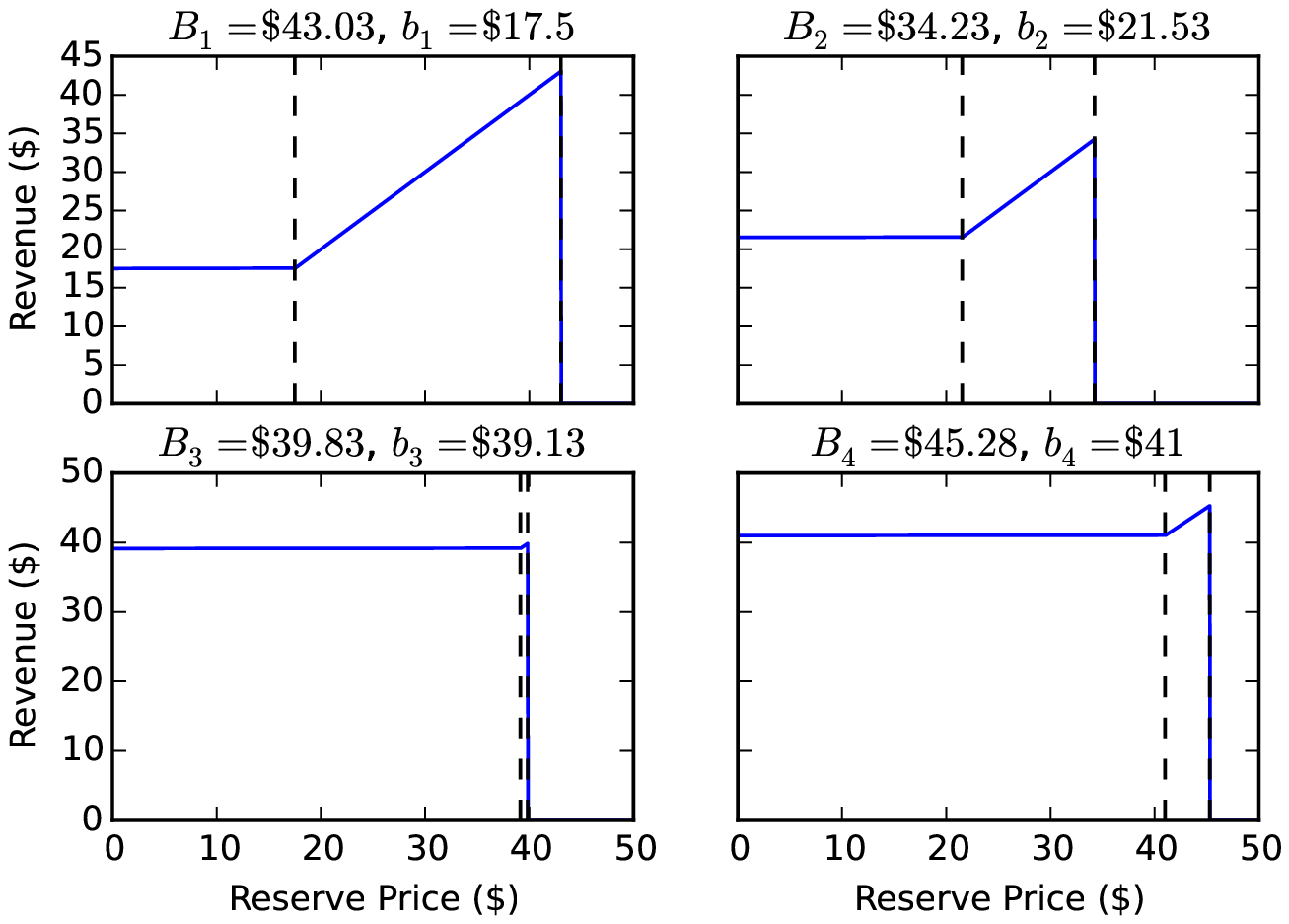}}
\subcaptionbox
{The effect of smoothing on the revenue function of an auction
  from the eBay data set. The smaller $\sigma$ the closer the smoothed
  revenue approximates the actual revenue function.\label{fig:smooth} }[0.49\textwidth]
{\includegraphics[width=0.49\textwidth]{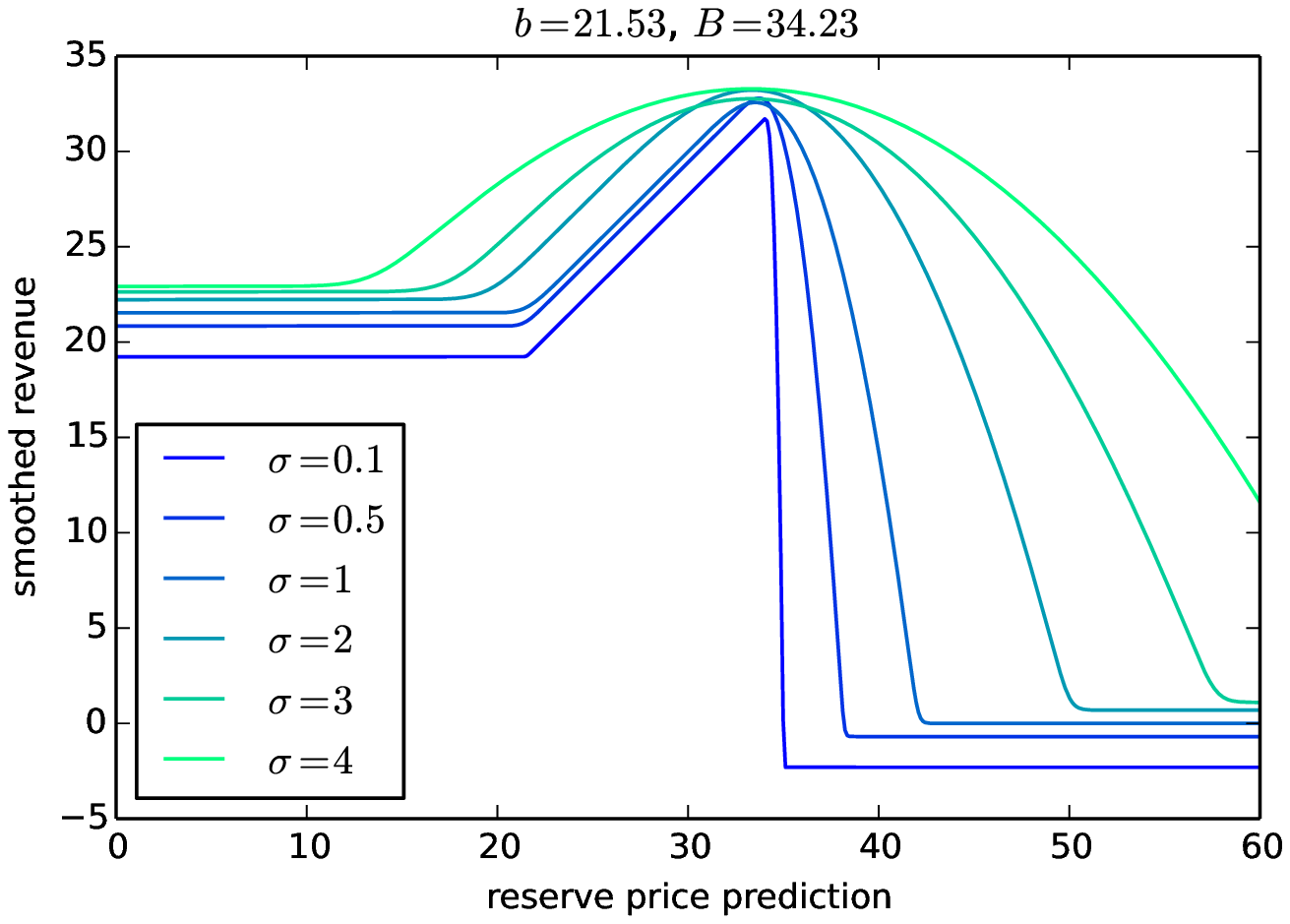}}
\caption{The revenue (a) and smoothed revenue (b) for example auctions from the eBay data set.}
\end{figure}

A typical solution to such real-valued prediction problems is linear
regression.  However, the solution to this problem is more
delicate. The reason is that the revenue function for each
auction---the amount of money that we make as a function of the
reserve price $y$---is asymmetric. It remains constant to the
second-highest bid $b$, increases to the highest bid $B$, and is zero beyond the highest bid. Formally,
\begin{align}
  \label{eq:revenue}
  R(y, B,b) = \begin{cases}
    b & \text{    if    } y <b\\
    y & \text{    if    } b\leq y \leq B\\
    0   & \text{    otherwise}
  \end{cases}.
\end{align}
Fig. \ref{revFun} illustrates this function for four auctions of sports
collectibles from eBay.
This figure puts the delicacy into
relief. The best reserve price, in retrospect, is the highest bid
$B$. But using a regression to predict the reserve price, e.g., by
using the highest bid as the response variable, neglects the important
fact that overestimating the reserve price is much worse than
underestimating it.  For example, consider the top left panel in
Fig. \ref{revFun}, which might be the price of a Stanley Kofax baseball card.
(Our data are anonymized, but we we use this example for
concreteness.)  The best reserve price in retrospect is \$43.03. 
A linear regressor is just as likely to overestimate as to underestimate and hence fails to reflect that setting the price in advance to \$44.00 would yield zero
earnings while setting it to \$40.00 would yield the full reserve.

To solve this problem we develop a new idea, \textit{the objective
  variable}.  Objective variables use the machinery of probabilistic
models to reason about difficult prediction problems, such as one that
seeks to optimize Eq.\ref{eq:revenue}.  Specifically, objective
variables enable us to formulate probabilistic models for which MAP
estimation directly uncovers profitable decision-making strategies.
We develop and study this technique to set the reserve price in
second-price auctions.

In more detail, our aim is to find a parameterized mechanism
$f(x_i; w)$ to set the reserve price from the auction features $x_i$.
In our study, we will consider a linear predictor, kernelized
regression, and a neural network.  We observe a historical data set of
$N$ auctions that contains features $x_i$, and the auction's two
highest bids $B_i$ and $b_i$; we would like to learn a good mechanism
by optimizing the parameter $w$ to maximize the total (retrospective)
revenue $\sum_{i=1}^N R(f(x_i; w), B_i, b_i)$.

We solve this optimization problem by turning it into a maximum a
posteriori (MAP) problem.  For each auction we define new binary
variables---these are the objective variables---that are conditional
on a reserve price. The probability of the objective variable being on
(i.e., equal to one) is related to the revenue obtained from the
reserve price; it is more likely on if the auction produces more
revenue.  We then set up a model that first assumes each reserve price
is drawn from the parameterized mechanism $f(x_i; w)$ and then draws
the corresponding objective variable.  Note that this model is defined
conditioned on our data, the features and the bids.  It is a model of
the objective variables.

With the model defined, we now imagine a ``data set'' where all of the
objective variables are on, and then fit the parameters $w$ subject to
these data.  Because of how we defined the objective variables, the
model will prefer more profitable settings of the parameters.  With
this set up, fitting the parameters by MAP estimation is equivalent to
finding the parameters that maximize revenue.

The spirit of this technique is that the objective variables are
likely to be on when we make good decisions, that is, when we profit
from our setting of the reserve price.  When we imagine that they are
all on, we are imagining that we made good decisions (in retrospect).
When we fit the parameters to these data, we are using MAP estimation
to find a mechanism that helps us make such decisions.

We first derive our method for linear predictors of reserve price and
show how to use the expectation-maximization
algorithm~\citep{dempster1977maximum} to solve our MAP problem.  We
then show how to generalize the approach to nonlinear predictors,
such as kernel regression and neural networks.  Finally, on simulated
data and real-world data from eBay, we show that this approach
outperforms the existing methods for setting the reserve price.  It is
both more profitable and more easily scales to larger data sets.

\parhead{Related work.}  Second-price auctions with reserve are first
introduced in \cite{easley}.  Ref.~\cite{ostrovsky2011reserve}
empirically demonstrates the importance of optimizing reserve prices;
Their study quantifies the positive impact it had on Yahoo!'s revenue.
However, most previous work on optimizing the reserve price are
limited in that they do not consider features of the
auction~\cite{ostrovsky2011reserve,cesa2013regret}.

Our work builds on the ideas in Ref.~\cite{mohri2014learning}.  This
research shows how to learn a linear mapping from auction features to
reserve prices, and demonstrates that we can increase profit when we
incorporate features into the reserve-price setting mechanism.  We
take a probabilistic perspective on this problem, and show how to
incorporate nonlinear predictors.  We show in Sec. \ref{sec:experiments} that
our algorithms scale better and perform better than these approaches.

The objective variable framework also relates to recent ideas from
reinforcement learning to solve partially observable Markov decision
processes (POMDPs)
\cite{toussaint2006probabilistic,toussaint2008hierarchical}.  Solving
an POMDP amounts to finding an action policy that maximizes expected
future return.
Refs.~\cite{toussaint2006probabilistic,toussaint2008hierarchical}
introduce a binary reward variable (similar to an objective variable)
and use maximum likelihood estimation to find such a policy.  Our work
solves a different problem with similar ideas, but there are also
differences between the methods.  In one way, the problem in
reinforcement learning is more difficult because the reward is itself
a function of the learned policy; in auctions, the revenue function is
known and fixed.  In addition, the work in reinforcement learning focuses
on simple discrete policies while we show how to use these ideas for
continuously parameterized predictors.

\section{Objective Variables for Second-Price Auctions with Reserve}\label{sec:SPAWR}

We first describe the problem setting and the objective.  Our data
come from previous auctions.  For each auction, we observe features
$x_i$, the highest bid $B_i$, and the second highest bid $b_i$.  The
features represent various characteristics of the auction, such as the
date, time of day, or properties of the item.  For example, one of the
auctions in the eBay sport collectibles data set might be for a Stanley Kofax baseball card; its
features include the date of the auction and various aspects of the
item, such as its condition and the average price of such cards on the
open market.

When we execute an auction we set a reserve price before seeing the
bids; this determines the revenue we receive after the bids are in.
The revenue function (Eq. \ref{eq:revenue}), which is indexed by the bids, determines how
much money we make as a function of the chosen reserve price. 
We illustrate this function for 4 auctions
from eBay in Fig. \ref{revFun}. Our goal is to use the historical data to
learn how to profitably set the reserve price from auction features, that
is, before we see the bids.

For now we will use a linear function to map auction features to a
good reserve price.  Given the feature vector $x_i$, we set the
reserve price with $f(x_i;w) = w^\top x_i$.  (In Sec. \ref{nonlin} we
consider nonlinear alternatives.)  We fit the coefficients $w$ from
data, seeking $w$ that maximizes the regularized revenue
\begin{align}
  \label{eq:regularized-revenue}
  w^* = \argmax \sum_{i=1}^N R(f(x_i;w), B_i, b_i) +
  (\lambda/2)w^\top w.
\end{align}
We have chosen an $L_2$ regularization controlled by parameter $\lambda$; other regularizers are also possible.

Before we discuss our solution to this optimization, we make two
related notes.  First, the previous reserve prices are \textit{not}
included in the data.  Rather, our data tell us about the relationship
between features and bids.  All the information about how much we
might profit from the auction is in the revenue function; the way
previous sellers set the reserve prices is not relevant.  Second, our
goal is not the same as learning a mapping from features to the
highest bid.  Not all auctions are made equal: Consider the top left auction in Fig. \ref{revFun} with highest and second highest bid $B_1 = \$43.03$ and $b_1 = \$17.5$ compared to the bottom left auction in Fig. \ref{revFun} with both highest and second highest bids almost identical at $B_3 = \$39.83$ and $b_3=\$39.17$. The profit margin in the first auction is much larger, so predicting the reserve price for this auction well is much more important than when the two highest bids are close to each other. We account for this by directly maximizing revenue,
rather than by modeling the highest bid.

\subsection{The smoothed revenue}

The optimization problem in Eq. \ref{eq:regularized-revenue} is difficult
to solve because $R(\cdot)$ is discontinuous (and thus non-convex).
Previous work~\cite{mohri2014learning} addresses
this problem by iteratively fitting differences of convex (DC) surrogate functions and solving the resulting DC-program \cite{tao1998dc}.
We define an objective function related to the revenue, but that smooths
out the troublesome discontunuity. In the next section we show how to
optimize this objective with an expectation-maximization algorithm.

We first place a Gaussian distribution on the reserve price centered
around the linear mapping, $y_i \sim \mathcal{N}(f(x_i; w),\sigma^2)$.
We define the smoothed regularized revenue to be
\begin{align}
  \label{eq:smoothed_revenue}
  \mathcal{L}(w) = \sum_{i=1}^N \log
  \mathbb{E}_{y_i}\left[\exp\left\{-R(y_i, B_i, b_i)\right\}\right] - (\lambda / 2) w^\top w.
\end{align}
Figure~\ref{fig:smooth} shows one term from Eq. \ref{eq:smoothed_revenue} and how -- for a specific auction -- the smoothed revenue becomes closer
to the original revenue function as $\sigma^2$ decreases.  
This approach was inspired by probit regression, where a Gaussian expectation is
introduced to smooth the discontinuous 0-1 loss% (i.e., classification loss)
~\citep{albert1993bayesian,holmes2006bayesian}.

We now have a well-defined and continuous objective function; in
principle, we can use gradient methods to fit the parameters.
However, we will fit them by recasting the problem as a regularized likelihood
under a latent variable model and then using the
expectation-maximization (EM) algorithm~\citep{dempster1977maximum}.
This leads to closed-form updates in both the E and M steps, and
facilitates replacing linear regression with a nonlinear predictor.

\subsection{Objective variables}
\label{sec:TOV}

To reformulate our optimization problem, we introduce the idea of the
the \textit{objective variable}.  Objective variables are part of a
probabilistic model for which MAP estimation recovers the parameter
$w$ that maximizes the smoothed revenue in Eq. \ref{eq:smoothed_revenue}.
Specifically, we define binary variables $z_i$ for each auction, each
conditioned on the reserve price $y_i$, the highest bid $B_i$, and
next bid $b_i$.  We can interpret these variables to indicate ``Is the
auction host satisfied with the outcome?''  Concretely, the likelihood
of satisfaction is related to how profitable the auction was relative
to the maximum profit, $p(z_i=1 \g y_i, B_i, b_i) = \pi(y_i, B_i, b_i)$
where
\begin{align}
  \label{eq:objective_probability}
  \pi(y_i, B_i, b_i) = \exp \left\{- (B_i - R(y_i, B_i,
  b_i))\right\}.
\end{align}
The revenue function $R(\cdot)$ is in Eq. \ref{eq:revenue}.  The revenue is
bounded by $B_i$; thus the probability is in $(0,1]$.

\begin{figure}[tp]
  \centering
\subcaptionbox
  {The objective variable model (OV model). The objective
    variable is shaded with diagonal lines to 
    distinguish that its value is not
    observed but rather set to our desired value.\label{fig:gm} }[0.5\textwidth]
{
  \begin{tikzpicture}
    \tikzstyle{main}=[circle, minimum size = 6mm, thick, draw =black!80,
    node distance = 8mm and 16mm] \tikzstyle{connect}=[-latex, thick]
    \tikzstyle{box}=[rectangle, draw=black!100]
    \node[main, fill = blue!20,pattern=north east lines] (z) [label=above:$z_i$] { };
    \node[main, fill = white!10] (y) [left=of z,label=above:$y_i$] {};
    \node[main] (w) [left=of y,label=below:$w$] { };
    \node[main, fill = black!40] (b) [below=of z,label=below:$B_i b_i$] {};
    \node[main, fill = black!40] (x) [left=of b,label=below:$x_i$] { };
    \path (x) edge [connect] (y)
    (w) edge [connect] (y)
    (b) edge [connect] (z)
    (y) edge [connect] (z);
    \node[rectangle, inner sep=0.6mm, fit= (b),label=right:$N$, yshift=-8mm, xshift=1mm] {};
    \node[rectangle, inner sep=8mm, draw=black!100, fit =(b) (x) (y) (z)] {};
  \end{tikzpicture}
}
\subcaptionbox
{For fixed w the posterior of the latent reserve price (red) is proportional to the prior (blue) times the likelihood of the objective (green). MAP estimation uncovers profitable modes of the posterior.\label{fig:posterior}}[0.49\textwidth]
{\includegraphics[width=0.45\textwidth]{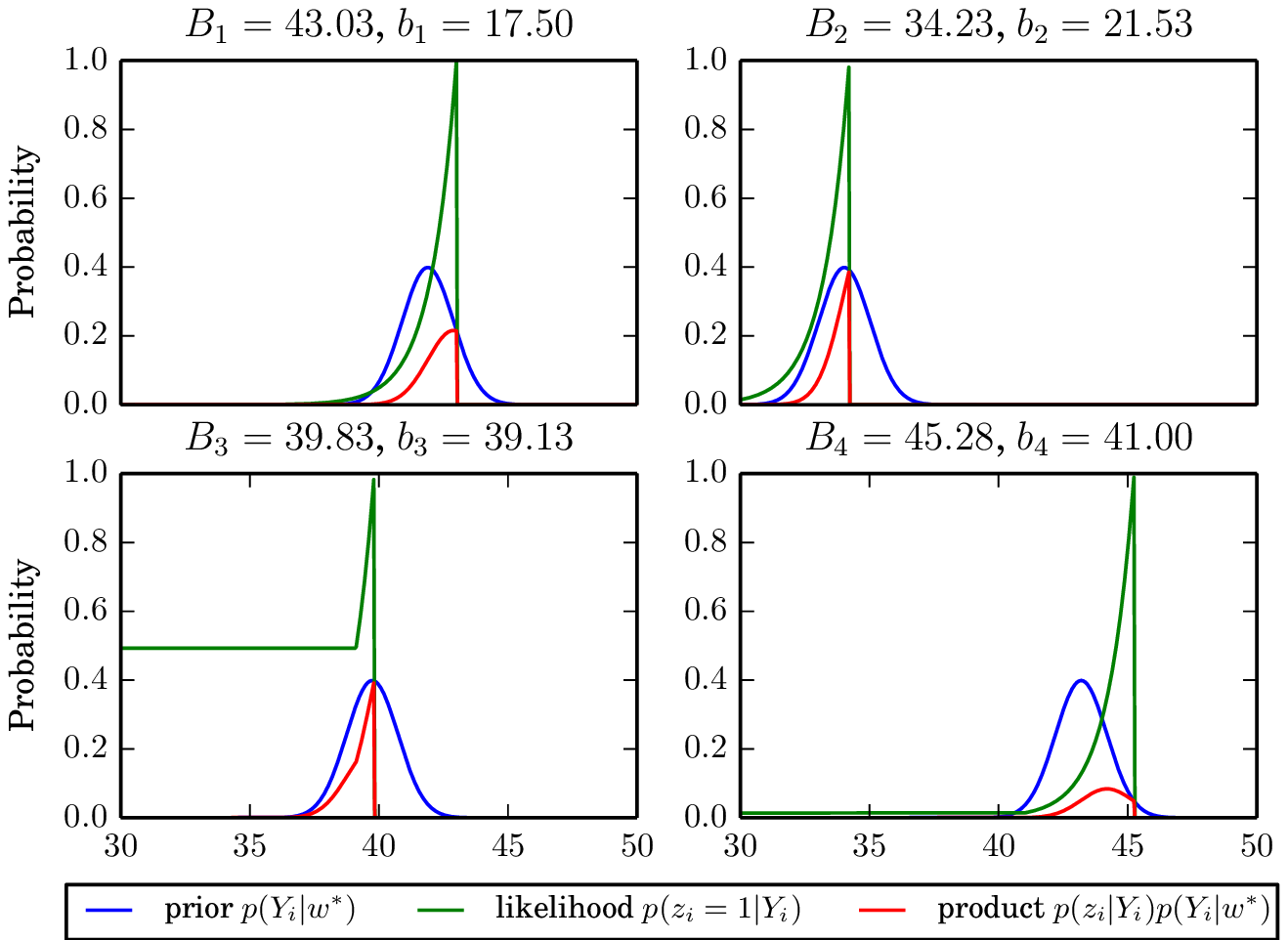}}
\caption{The objective variable framework transforms the revenue maximization task into a MAP estimation task. The model and the hallucinated data are designed such that the modes of the model's posterior are the local maxima of the smoothed revenue in Eq. \ref{eq:smoothed_revenue}}.
\end{figure}

What we will do is set up a probability model around the objective
variables, assume that they are all ``observed'' to be equal to one
(i.e., we are satisfied with all of our auction outcomes), and then
fit the parameter $w$ to maximize the posterior conditioned on this ''hallucinated data''.
Fig. \ref{fig:posterior} provides visual intuition why the modes of the posterior are profitable.
For fixed $w$ the posterior of $y_i$ is proportional to the product of its prior centered at $f(x_i;w)$ and the likelihood of the objective variable (Eq. \ref{eq:objective_probability}) which captures the profitability of each possible reserve price prediction.

Consider the following model,
\begin{align}\label{eqn:generative}
  w & \sim \mathcal{N}(0, \lambda^{-1} I) \\
  y_i \g w, x_i & \sim \mathcal{N}(f(x_i;w), \sigma^2) \quad i \in \{1,
        \ldots, N\} \\
  z_i \g y_i, B_i, b_i & \sim \mathrm{Bernoulli}(\pi(y_i, B_i, b_i))
\end{align}
where $f(x_i;w)=x_i ^ \top w$ is a linear map (for now). This is illustrated as a graphical model in Fig. \ref{gm}.

Now consider a data set $\bz$ where all of the objective variables $z_i$ are
equal to one.  Conditional on this data, the log posterior of $w$
marginalizes out the latent reserve prices $y_i$,
\begin{align}
  \log p(w \g \bz, \bx, \bB, \bb) =
  \log p(w \g \lambda) + \sum_{i=1}^{N} \left(\log \E\left[\exp\{R(y_i, B_i,
  b_i)\}\right] - B_i\right) - C,
\end{align}
where $C$ is the normalizer.
This is the smoothed revenue of Eq. \ref{eq:smoothed_revenue} plus a constant involving
the top bids $B_i$ in Eq. \ref{eq:objective_probability}, constant
components of the prior on $w$, and the normalizer.  
Thus, we can optimize the smoothed revenue by taking MAP estimates of $w$.

As we mentioned above, we have defined variables corresponding to the
auction host's satisfaction.  With historical data of auction
attributes and bids, we imagine that the host was satisfied with every
auction.  When we fit $w$, we ask for the reserve-price-setting
mechanism that leads to such an outcome.

\subsection{MAP estimation with expectation-maximization}
\label{sec:EM}

The EM algorithm is a technique for maximum likelihood estimation in
the face of hidden variables~\citep{dempster1977maximum}.  (When there
are regularizers, it is a technique for MAP estimation.)  In the
E-step, we compute the posterior distribution of the hidden variables
given the current model settings; in the M-step, we maximize the
expected complete regularized log likelihood, where the expectation is
taken with respect to the previously computed posterior.

In the OV model, the latent variables are the reserve prices $\by$;
the observations are the objective variables $\bz$; and the model
parameters are the coefficients $w$.  We compute the posterior
expectation of the latent reserve prices in the E-step and fit the
model parameters in the M-step.
This is a coordinate ascent algorithm on the expected complete regularized 
log likelihood of the model and the data.
Each E-step tightens the bound on the likelihood and the new bound is then optimized in the M-step.

\parhead{E-step.} At iteration $t$, the E-step computes the
conditional distribution of the latent reserve prices $y_i$ given the
objective variables $z_i = 1$ and the parameters
$w^{(t-1)}$ of the previous iteration.  It is
\begin{align}
  \label{qYi}
  p(y_i|z_i=1,w^{(t-1)})
  & \propto p(z_i=1|y_i) p(y_i|w^{(t-1)}) \\
  & \propto \exp\left\{-(B_i-R(y_i, B_i, b_i)\right\}
            \phi\left(\frac{y_i-f(x_i; w^{(t-1)})}{\sigma}\right).
\end{align}
where $\phi(\cdot)$ is the pdf of the standard normal distribution.
The normalizing constant is in the appendix in Eq. \ref{normalizing}; we
compute it by integrating Eq. \ref{qYi} over the real line. We can then
compute the posterior expectation
$\E\left[y_i \g z_i, w^{(t-1)}\right]$ by using the moment generating
function. (See Eq. \ref{EY},Sec. \ref{sec:updateEY})

\parhead{M-step.}  The M-step maximizes the complete joint log-likelihood
with respect to the model parameters $w$.  When we use a linear predictor to set the reserve prices, i.e. $f(x_i;w) = x_i^\top w$, the M-step has a closed form
update, which amounts to ridge regression against response variables
$\E\left[y_i \g z_i, w^{(t-1)}\right]$ (Eq. \ref{EY}) computed in the
E-step. The update is
\begin{align}
  w^{(t)} = \left(\lambda I + \frac{1}{\sigma^2}\bx^\top
  \bx\right)^{-1} \frac{1}{\sigma^2} \bx^\top \E\left[\by \g
  \bz, w^{(t-1)}\right]
  \label{wls}
\end{align}
where $\E\left[\by \g \bz, w^{(t-1)}\right]$ denotes the vector with
$i^{th}$ entry $\mathbb{E}\left[y_i \g \bz, w^{(t-1)}\right]$ and
similarly $\bx$ is a matrix of all feature vectors $x_i$.

\parhead{Algorithm details.} To initialize, we set the expected
reserve prices to be the highest bids $\E[y_i \g z_i] = B_i$ and run
an M-step.  The algorithm then alternates between updating the weights
using Eq. \ref{wls} in the M-step and then integrating out the latent
reserve prices in the E-step. The algorithm terminates when the change
in revenue on a validation set is below a threshold. (We use $10^{-5}$.)

The E-step is linear in the number of auctions $N$ and can be parallelized since the expected reserve prices are conditionally independent in our model.
The  least squares update has asymptotic complexity $O(d^2N)$ where $d$ is the number of features.

\subsection{Nonlinear Objective Variable Models}
\label{sec:nonlin}

One of the advantages of our EM algorithm is that we can change the
parameterized prediction technique $f(x_i; w)$ from which we map
auction features to the mean of the reserve price. So far we have only
considered linear predictors; here we show how we can adapt the
algorithm to nonlinear predictors. As we will see in
Sec. \ref{sec:experiments}, these nonlinear predictors outperform the
linear predictors.

In our framework, much of the model in Fig. \ref{fig:gm} and corresponding
algorithm remains the same even when considering nonlinear predictors.
The distribution of the objective variables is unchanged
(Eq. \ref{eq:objective_probability}) as well as the E-step update in the
EM algorithm (Eq. \ref{EY}). All of the changes are in the M-step.

\parhead{Kernel regression.} Kernel
regression~\cite{aizerman1964theoretical} maps the features $x_i$ into
a higher dimensional space through feature map $\psi(\cdot)$; the mechanism for
setting the reserve price becomes $f(x_i;w) = \psi(x_i)^Tw$. In kernel
regression we work with the $N \times N$ Gram matrix $K$ of inner
products, where $K_{ij} = \psi(x_i)^T \psi(x_j)$. In this work we use
a polynomial kernel of degree $D$, and thus compute the gram matrix
without evaluating the feature map $\psi(\cdot)$ explicitly, $K = (\bx^\top \bx + 1)^D$.

Rather than learning the weights directly, kernel methods operate in
the dual space $\alpha \in \mathbb{R}^N$.  If $K_i$ is the $i^{th}$
column of the Gram matrix, then the mean of the reserve price is
\begin{align}
  f(x_i;w) = \psi(x_i)^Tw=K_i^T\alpha.
\end{align}
The corresponding M-step in the algorithm becomes
\begin{align}
  \label{eqn:kls}
  \alpha^{(t)} = \left(\frac{1}{\sigma^2} K +\lambda
  I_N\right)^{-1}\frac{1}{\sigma^2}\mathbb{E}[\by \g \bz, \alpha^{(t-1)}].
\end{align}
See \cite{bishop2006pattern} for the technical details around kernel regression.

We will demonstrate in Sec. \ref{sec:experiments} that replacing linear
regression with kernel regression can lead to better reserve price
predictions. However, working with the Gram matrices comes at a computational cost and we consider neural networks as a scalable alternative to
infusing nonlinearity into the model.

\parhead{Neural networks.} We also explore an objective variable model
that uses a neural network~\cite{bishop1995neural} to set the mean
reserve prices. We use a network with one hidden layer of $H$ units
and activation function $\tanh(\cdot)$. The parameters of the neural
net are the weights of the first layer and the second layer:
$w = \{w^{(1)} \in \mathbb{R}^{H \times d}, w^{(2)} \in
\mathbb{R}^{1\times H}\}$.
The mean of the reserve price is
\begin{align}\label{eqn:nn}
  f(x_i;w) = w^{(2)}(\tanh(w^{(1)}x_i)).
\end{align}
The M-step is no longer analytic; Instead, the network is trained using stochastic gradient
methods.

\section{Empirical Study}
\label{sec:experiments}

\begin{table}[t]
\centering
\footnotesize
\caption{The performance of the EM algorithms from Sec. \ref{sec:SPAWR}
  (OV Regression, OV Kernel Regression with degree $2$ and $4$, OV Neural Networks)
  against the current state of the art (DC~\cite{mohri2014learning}
  and NoF~\cite{cesa2013regret}).  We report results in terms of
  percentage of maximum possible revenue (computed by an
  oracle that knows the highest bid in advance).  For each data set, we report mean and
  standard error aggregated from ten train/validation/test splits.
  Our methods outperform the existing methods on all data.}
\label{tab:results}
\begin{tabular}{| c || c | c | c | c || c | c | } \hline
                     &OV Reg        & OV Kern (2)    & OV
                                                                Kern (4)
                                                                 &
                                                                       OV
                                                                       NN
  &DC~\cite{mohri2014learning} & NoF~\cite{cesa2013regret} \\
\hline\hline
Linear Sim.           &$\mathbf{81.4\pm0.2}$&$81.2\pm0.2$&$78.2\pm0.6$&$72.2\pm1.5$&$80.3\pm0.3$&$49.9\pm0.1$\\
\hline
Nonlinear Sim.           &$50.3\pm0.3$&$66.2\pm0.4$&$\mathbf{70.1\pm0.6}$&$63.7\pm2.9$&$59.4\pm2.0$&$49.9\pm0.2$\\
\hline
eBay (s)              &$61.0\pm0.7$&$63.7\pm3.0$&$63.4\pm2.8$&$\mathbf{74.4\pm1.1}$&$59.5\pm1.1$&$55.8\pm0.3$\\
\hline
eBay (L)              &$62.4\pm0.2$&      -     &     -      &$\mathbf{84.0\pm0.2}$&      -     &$56.0\pm0.1$\\
\hline
\end{tabular}
\end{table}

We studied our algorithms with two simulated data sets and a large
collection of real-world auction data from eBay.  In each study, we
fit a model on a subset of the data (using a validation set to set
hyperparameters) and then test how profitable we would be if we used
the fitted model to set reserve prices in a held out set.  Our
objective variable methods outperformed the existing state of the art.

\parhead{Data sets and replications.} We evaluated our method on both
simulated data and real-world data.
\begin{itemize}[leftmargin=*]

\item \textit{Linear simulated data.} Our simplest
  simulated data contains $d=5$ auction features. We drew
  features
  $x_i \sim N(0,I) \in \mathbb{R}^d$ for 2,000 auctions; we drew a
  ground truth weight vector $\hat{w} \sim \cN(0,I) \in
  \mathbb{R}^{d}$
  and an intercept $\alpha \sim \cN(0,1)$; we drew the highest
  bids for each auction from the regression
  $B_i \sim \mathcal{N}(w^\top x_i + \alpha, 0.1)$ and set the second
  bids $b_i = B_i / 2$. (Data for which $B_i$
  is negative are discarded and re-drawn.) We split into
  $N_{\text{train}} = 1000$ and 
  $N_{\text{valid}} = N_{\text{test}}=500$.

\item \textit{Nonlinear simulated data.} These data contain features
  $x_i$, true coefficients $\hat{w}$, and intercept $\alpha$ generated
  as for the linear data. We generate highest bids by taking the
  absolute value of those generated by the regression and second
  highest bids by halving them, as above. Taking
  the absolute value introduces a nonlinear relationship between
  features and bids.

\item \textit{Data from eBay.} Our real-world data is auctions of 
  sports collectibles from 
  eBay.\footnote{This data set comes from http://cims.nyu.edu/~munoz/data/index.html}
  There are $d=74$
  features. All
  covariates are centered and rescaled to have mean zero and standard
  deviation one. We analyze two data sets from eBay, one small and one
  large. On
  the small data
  set, the total number of auctions
  is $6,000$, split into $N_{\text{train}} =  N_{\text{valid}}
  = N_{\text{test}}=2,000$. On the large data set the total number is
  70,000, split into $N_{\text{train}} = 50,000$, and 
  $N_{\text{valid}} =N_{\text{test}} = 10,000$. 
\end{itemize}

In our study, we fit each method on the training data, use the
validation set to decide on hyperparameters, and then evaluate the
fitted predictor on the test set, i.e., compute how much revenue we
make when we use it to set reserve prices. For each data set, we
replicate each study ten times, each time randomly creating the
training set, test set, and validation set.

\parhead{Algorithms.}  We describe the objective variable
algorithms from Sec. \ref{sec:SPAWR}, all of which we
implemented in Theano~\citep{bergstra2010scipy,bastien2012theano}, as well as the two previous methods we compare against.

\begin{itemize}[leftmargin=*]

\item \textit{OV Regression}. OV Regression learns a linear predictor
  $w$ for reserve prices using the algorithm in Sec. \ref{sec:EM}. We find a
  good setting for the smoothing parameter $\sigma$ and regularization
  parameter $\lambda$ using grid search.

\item \textit{OV Kernel Regression.} OV Kernel Regression uses a
  polynomial kernel to predict the mean of the reserve price; we study
  polynomial kernels of degree 2 and 4.

\item \textit{OV Neural Network.} OV Neural Network fits a neural net
  for predicting the reserve prices. As we discussed in
  Sec. \ref{sec:nonlin}, the M-step uses gradient optimization; we used
  stochastic gradient ascent with a constant learning rate and early
  stopping~\cite{prechelt2012early}.  Further, we used a warm-start
  approach, where the next M-step is initialized with the results of
  the previous M-step. We set the number of hidden units to $H = 5$
  for the simulated data and $H=100$ for the eBay data. We use grid
  search to set the smoothing parameter $\sigma$, the regularization
  parameters, the learning rate, the batch size, and the number of
  passes over the data for each M-step.

\item \textit{Difference of Convex Functions
    (DC)~\cite{mohri2014learning}.} The DC algorithm finds a linear
  predictor of reserve price with an iterative procedure based on
  DC-programming~\cite{tao1998dc}. Grid search is used on the
  regularization parameter as well as the margin to select the
  surrogates for the auction loss.

\item \textit{No Features (NoF)~\citep{cesa2013regret}.} This is the
  state-of-the-art approach to set the reserve prices when we do not
  consider the auction's features. The algorithm iterates over the
  highest bids in the training set and evaluates the profitability of
  setting all reserve prices to this value on the training set.
  Ref.~\cite{mohri2014learning} gives a more efficient algorithm based
  on sorting.

\end{itemize}

\parhead{Results.}  Tab. \ref{tab:results} gives the results of our study.
The metric is the percentage of the highest possible revenue, where an
oracle anticipates the bids and sets the reserve price to the highest
bid.

A trivial strategy (not reported) sets all reserve prices to zero, and
thus earns the second highest bid on each auction.  The algorithm
using no features~\cite{cesa2013regret} does slightly better than this
but not as well as the algorithms which use features.  OV Regression
[this paper] and DC~\cite{mohri2014learning} both fit linear mappings
and exhibit similar performance.  However, the DC algorithm does not
scale to the large eBay data set.

The nonlinear OV algorithms (OV Kernel Regression and OV Neural
Networks) outperform the linear models on the nonlinear simulated data
and the real-world data.  Note that the kernel algorithms do not scale
to the large eBay data set because working with the Gram matrix
becomes infeasible as the training set gets large.  OV Neural Networks
significantly outperforms the existing methods on the real-world data.
This is a viable solution to maximizing profit from historical auction
data.
\section{Summary and Discussion}
We developed the objective variable framework for combining
probabilistic modeling with optimal decision making.  We used this
method to solve the problem of how to set the reserve price in
second-price auctions.  Our algorithms scaled better and outperformed
the current state of the art on both simulated and real-world data.
%\clearpage
\appendix
\section{Appendix - Update Equations for EM}\label{sec:updateEY}
The normalizing constant $C_i$ of Eq. \ref{qYi} can be computed by integrating Eq. \ref{qYi} over the real line. 
Let $\mu_i = f(x_i;w^{(t-1)})$. Up to a constant factor of $e^{B_i}$ the normalizing constant $C_i$ then equals
\begin{align}\label{normalizing}
C_ie^{B_i} =&  \int_{-\infty}^{b_i}e^{b_i} \phi(\frac{y_i-\mu_i}{\sigma})dy_i + \int_{b_i}^{B_i}e^{y_i} \phi(\frac{y_i-\mu_i}{\sigma})
 + \int_{B_i}^{\infty} \phi(\frac{y_i-\mu_i}{\sigma})\\
 =&\sigma e^{b_i}\Phi(\frac{b_i-\mu_i}{\sigma}) 
+ \sigma e^{\mu_i+\frac{\sigma^2}{2}}\big[\Phi(\frac{B_i-(\mu_i+\sigma^2)}{\sigma})-\Phi(\frac{b_i-(\mu_i+\sigma^2)}{\sigma})\big] \nonumber \\
&+ \sigma \big[ 1 - \Phi(\frac{B_i-\mu_i}{\sigma})\big] . 
\end{align}

Computing the expectation of the latent reserve price $\mathbb{E}[y_i]$ entails evaluating the moment generating function $M_i(s)=\mathbb{E}[e^{sy_i}]$, where expectation is taken w.r.t. the posterior $p(y_i|z_i=1,w^{(t-1)})$. 
Taking the derivative with respect to $s$ and setting $s=0$ then yields the desired expectation.
\begin{align}
\mathbb{E}[y_i] = & \frac{dM_i(s)}{ds}\bigg|_{s=0}\\
 = & \frac{\sigma e^{b_i}}{C_i}\mu_i \Phi(\frac{b_i - \mu_i }{\sigma})- \frac{\sigma^2 e^{b_i}}{C_i} \phi(\frac{b_i - \mu_i }{\sigma})
          + \frac{\sigma}{C_i}(\mu_i + \sigma^2)e^{\mu_i+\frac{\sigma^2}{2}}\Phi(\frac{B_i - (\mu_i + \sigma^2)}{\sigma} )\nonumber\\
         & - \frac{\sigma}{C_i}(\mu_i + \sigma^2)e^{\mu_i+\frac{\sigma^2}{2}}\Phi(\frac{b_i- (\mu_i + \sigma^2)}{\sigma} )+ \frac{\sigma }{C_i}\mu_i \big[1 - \Phi(\frac{B_i - \mu_i }{\sigma} )\big]\nonumber\\
         & - \frac{\sigma^2}{C_i}e^{\mu_i+\frac{\sigma^2}{2}} \big[\phi(\frac{B_i - (\mu_i + \sigma^2)}{\sigma} )
-\phi(\frac{b_i- (\mu_i + \sigma^2)}{\sigma} )\big]
  + \frac{\sigma^2 }{C_i} \phi(\frac{- \mu_i }{\sigma} )\label{EY}
\end{align}
\clearpage
\bibliographystyle{unsrtnat}
\bibliography{myrefs}
\end{document}

%% file: preamble.tex
\usepackage{amssymb}

\usepackage{graphicx}                               % graphics
\usepackage{amsmath}                                % math
\usepackage[labelfont=bf]{caption}                  % bold figure/table label
\usepackage[format=hang]{subcaption}                % subfigures
\usepackage{booktabs}                               % nice tables
\usepackage[linktoc=all]{hyperref}
\usepackage[all]{hypcap}
\usepackage[numbers,sort]{natbib}                   % better bibliography
\usepackage[acronym,smallcaps,nowarn]{glossaries}   % second best package ever
\usepackage{enumitem}

\DeclareRobustCommand{\parhead}[1]{\textbf{#1} }
\usepackage{bbm}
\usepackage{algorithm}
\usepackage{algcompatible}
\usepackage{bbm}
\usepackage{tikz}
\usepackage{subcaption}
\usetikzlibrary{fit,positioning,patterns}

%% file: preamble_math.tex
\newcommand{\E}{\mathbb{E}}

\newcommand{\cN}{\mathcal{N}}

\DeclareMathOperator*{\argmax}{arg\,max}

\newcommand{\g}{\, | \,}

%% file: preamble_acronyms.tex
\newacronym{KL}{kl}{Kullback-Leibler}
\newacronym{ELBO}{elbo}{evidence lower bound}

\newacronym{SVI}{svi}{stochastic variational inference}

\newacronym{AD}{ad}{automatic differentiation}

\newacronym{VBSTAN}{vbstan}{variational Bayes in Stan}

\newacronym{ARD}{ard}{automatic relevance determination}

\newacronym{EP}{ep}{expectation propagation}
\newacronym{EPA}{epa}{expectation propagation algorithm}